\documentclass[letterpaper, 10 pt, conference]{ieeeconf}  

\IEEEoverridecommandlockouts                              

\usepackage{graphics} 
\usepackage{epsfig} 
\usepackage{mathptmx} 
\usepackage{times} 
\usepackage{amsmath} 
\usepackage{amssymb}  
\usepackage{siunitx}

\usepackage[dvipsnames]{xcolor}
\usepackage{multicol}
\usepackage[bookmarks=truem]{hyperref}
\hypersetup{
  colorlinks = true, 
  urlcolor = blue, 
  linkcolor = blue, 
  citecolor = Cerulean, 
}
\usepackage{comment}
\usepackage{xspace}
\usepackage{booktabs}
\usepackage{multirow}
\usepackage{cellspace, tabularx}
\usepackage{bbm}
\usepackage{csquotes}
\usepackage{gensymb}
\usepackage[ruled, lined, linesnumbered, commentsnumbered, longend]{algorithm2e}
\usepackage[hang,flushmargin]{footmisc}
\usepackage{lipsum}
\usepackage{tablefootnote} 
\usepackage{float}
\usepackage{wrapfig}
\usepackage{tablefootnote}
\usepackage{caption}
\usepackage{subcaption}

\usepackage[backend=biber,
            hyperref=true,
            url=false,
            isbn=false,
            doi=false,
            backref=false,
            style=ieee,
            citestyle=numeric-comp,
            sorting=nyt,
            block=none]{biblatex}
\usepackage[font=footnotesize]{caption}
\renewcommand{\bibfont}{\small}
\newcommand{\ourtool}{\textsc{KinScene}}

\makeatletter
\DeclareRobustCommand\onedot{\futurelet\@let@token\@onedot}
\def\@onedot{\ifx\@let@token.\else.\null\fi\xspace}

\def\eg{\emph{e.g}\onedot} 
\def\ie{\emph{i.e}\onedot}

\makeatother

\newcommand{\ssfont}[1]{#1}

\SetKwComment{Comment}{/* }{ */}

\SetCommentSty{mycommfont}

\makeatletter
\newcommand{\removelatexerror}{\let\@latex@error\@gobble}
\makeatother

\title{\LARGE \bf
\ourtool{}: Model-Based Mobile Manipulation of Articulated Scenes
}

\author{Cheng-Chun Hsu$^{1}$, Ben Abbatematteo$^{1}$, Zhenyu Jiang$^{1}$, Yuke Zhu$^{1}$, Roberto Mart{\'i}n-Mart{\'i}n$^{1*}$, Joydeep Biswas$^{1*}$\\
    \thanks{$^{1}$ Department of Computer Science, The University of Texas at Austin. $^{*}$~equal advising. Correspondence to {\tt\small chengchun@utexas.edu}.}
}

\begin{document}

\maketitle
\thispagestyle{empty}
\pagestyle{empty}


\begin{abstract}
Sequentially interacting with articulated objects is crucial for a mobile manipulator to operate effectively in everyday environments. To enable long-horizon tasks involving articulated objects, this study explores building scene-level articulation models for indoor scenes through autonomous exploration. 
While previous research has studied mobile manipulation with articulated objects by considering object kinematic constraints, it primarily focuses on individual-object scenarios and lacks extension to a scene-level context for task-level planning. To manipulate multiple object parts sequentially, the robot needs to reason about the resultant motion of each part and anticipate its impact on future actions.
We introduce \ourtool{}, a full-stack approach for long-horizon manipulation tasks with articulated objects. The robot maps the scene, detects and physically interacts with articulated objects, collects observations, and infers the articulation properties. For sequential tasks, the robot plans a feasible series of object interactions based on the inferred articulation model. We demonstrate that our approach repeatably constructs accurate scene-level kinematic and geometric models, enabling long-horizon mobile manipulation in a real-world scene. Code and additional results are available at \href{https://chengchunhsu.github.io/KinScene/}{\url{https://chengchunhsu.github.io/KinScene/}}
\end{abstract}
\section{Introduction}

\begin{figure}[t]
    \centering
    \includegraphics[width=\linewidth]{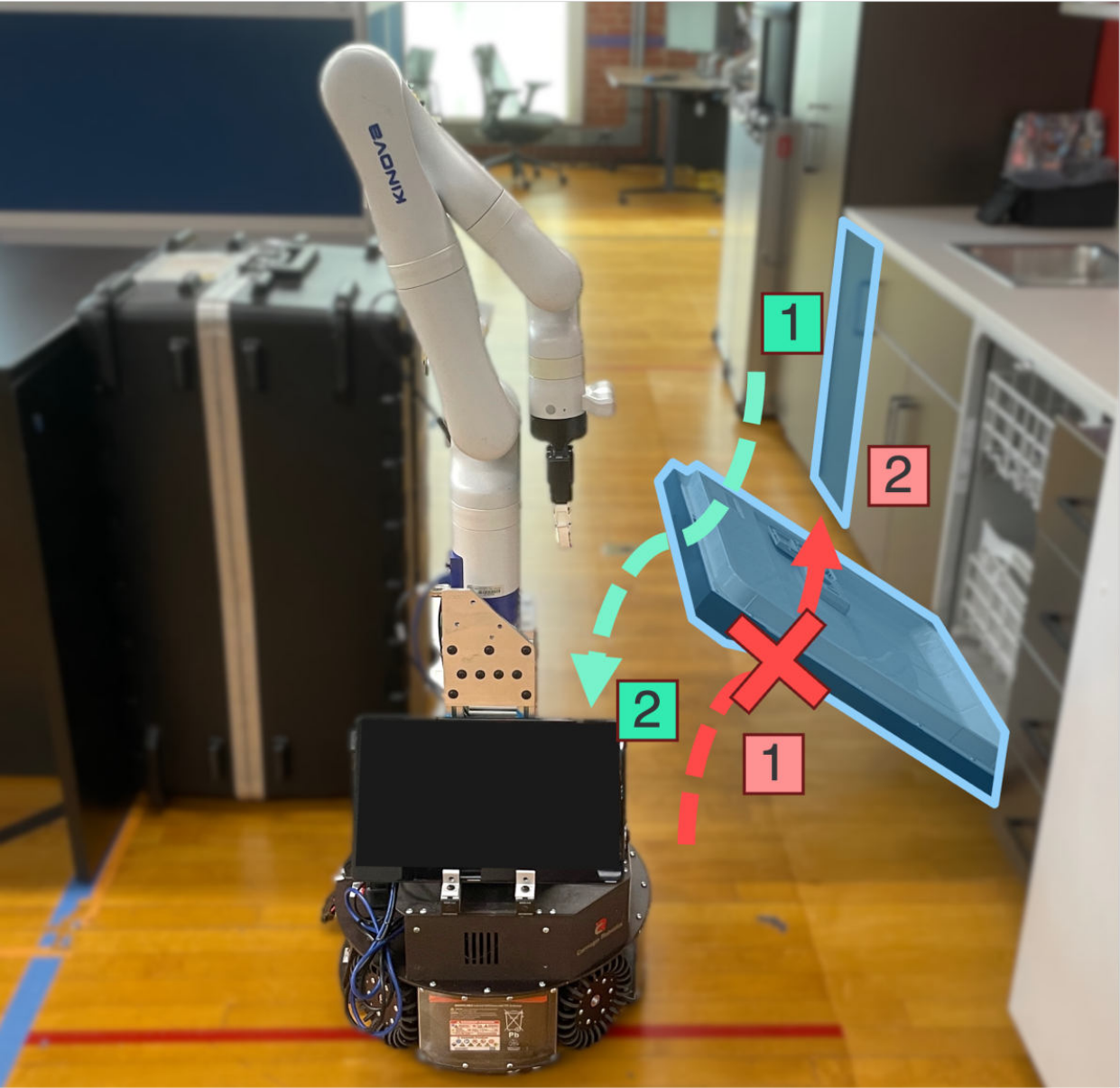}
    \caption{
        \textbf{\ourtool{} enables scene-level reasoning about articulated objects}. 
        In this scenario, attempting to open the dishwasher would obstruct the path for subsequent interactions. \ourtool{} constructs a scene-level articulation model and plans a feasible trajectory.  
    }
    \vspace{-3mm}
    \label{fig: teaser}
\end{figure}

Domestic robots in human unstructured environments need to perform long-horizon tasks by reasoning about and actuating
articulated objects at a \emph{scene-level}. For example, when putting away the dishes
from the dishwasher, the robot must consider actuating the dishwasher door and drawer, as well as the doors and drawers of the cabinet.
Reasoning solely about individual joints prevents the robot from understanding intra-object dependencies such as that the dishwasher door needs to be actuated before the dishwasher drawer, and scene-level dependencies such as that the dishwasher door may block the path to the cabinet (Fig.~\ref{fig: teaser}).

Previous work has investigated in isolation individual parts of the problem of building, planning and manipulating articulation at the scene level including single-joint estimation with passive sensing~\cite{sturm2011probabilistic, abbatematteo2019learning, li2020category}, zero-shot actuation of unknown objects~\cite{martin2022coupled, mo2021where2act, gadre2021act, wu2021vat}, and task and motion planning given a scene-level articulation model~\cite{stilman2007task,baum2017opening, garrett2021integrated}, but key open questions that persist are 1) \emph{how to integrate} these components, 2) what are the \emph{assumptions and limitations} inherent to each individual component for effective synthesis, and 3) \emph{how well} different solutions work when integrated in an end-to-end automated solution.

In this paper, we introduce \ourtool{}, a model-based solution to scene-level articulation reasoning and manipulation that involves autonomously exploring the scene to generate a geometric-kinematic model, and using this model to plan and execute scene-informed manipulation plans. \ourtool{} consists of three phases: First, in the \emph{mapping stage}, the robot builds a static map and identifies potential articulated objects. Next, in the \emph{articulation discovery stage}, the robot autonomously visits each articulated object and manipulates it to model its joint parameters and build a scene-level articulation model. Finally, in the \emph{scene-level manipulation stage}, the robot leverages the scene-level articulation model to efficiently plan and execute actions to manipulate articulated objects in the scene to complete long-horizon tasks.

We evaluate our approach in a real-world kitchen that features diverse everyday articulated objects of varying sizes and positions and compare \ourtool{} with several choices of individual articulation estimation components. Furthermore, we perform ablations showing that autonomous exploration is essential for articulated object manipulation. The experiments demonstrate that our robot can accurately infer articulation properties through exploration. Leveraging the scene-level articulation model, our robot enhances execution speed by 60.55\% and ultimately achieves a higher success in planning and manipulating the articulated objects by reasoning at a scene scale.

\section{Problem Formulation}
\label{sec: formulation}

Our goal is to change the kinematic state of the scene to a desired configuration. We assume this configuration is generated by a higher-level task planner and it is needed to perform a long-horizon task. For example, to \textit{unload the dishwasher} the robot needs to open the door of the dishwasher before picking the clean dishes, and open the cabinet doors before putting away the dishes. Our goal will be to achieve these states, \ie{}, planning and executing a sequence of single joint interactions to bring the environment to the desired configuration, taking into account the constraints introduced by the scene-level articulation. 

Formally, we represent the scene-level articulation model as a kinematic tree~\cite{sturm2011probabilistic}, $M = \langle B, V, E \rangle$, consisting of a static 3D base map, $B$, (\eg{} including the walls and fixed kitchen counters), vertices $V=\{v_i|i\in[1,N]\}$ consisting of $N$ 3D models of articulated moving parts, and articulated edges, $E=\{e_i|i\in[1,N]\}$, with kinematic constraints between each articulated moving part and the base map~\cite{hsu2023ditto}. Each edge, $e_i$, may be either a prismatic or a revolute joint, and consists of the associated joint parameters and articulation limits. Revolute joint parameters are the direction of the revolute axis $u^r \in \mathbb{R}^{3}$ and a pivot point $q \in \mathbb{R}^{3}$ on the revolute axis; prismatic joint parameters are the direction of the translation axis $u^p \in \mathbb{R}^{3}$. The scene-level kinematic state is thus represented by the state of every articulated edge, $\Theta=\{\theta_i|i\in[1,N]\}$. 
The goal will be then to change the kinematic state of the scene from an initial state, $\Theta_0$, to a given goal state, $\Theta_g$, that involves actuating several joints.
To solve this problem, our solution will plan and execute a sequence of single joint changes, $\Pi = [\theta_1, \ldots, \theta_T]$, such that the actuation of each $\theta_i$ is feasible and the goal scene-level configuration $\Theta_g$ is achieved. As the scene-level articulation model $M$ is unknown to the robot a priori, the robot must first explore the environment to reconstruct the scene kinematics and geometry from egocentric observations. This necessitates an \textit{articulation discovery stage} prior to deployment in which the robot gathers observations and constructs the model $M$. %

As the task involves a mobile robot actuating multiple articulated objects in the environment to achieve a desired configuration, we represent the state space as the configuration of both the robot and articulated objects. The state space, denoted $S$, is defined as $S = \Theta \times S_R$, where $\Theta$ represents the state for each joint in $E$ (articulated objects in the environment), and $S_R$ represents the state of the robot. The robot state $S_R = S_{base} \times S_{arm} \times S_{gripper}$ consists of the base pose $S_{base} \in SE(2)$, arm configuration $S_{arm} \in \mathcal{Q}$, and the gripper's openness $S_{gripper} \in \{0, 1\}$. 

To achieve the desired scene configuration, the robot must reason both at the task level (i.e. which objects to manipulate) and the motion level (i.e. finding collision-free motion plans). As such, we cast this problem as a hierarchical planning problem~\cite{garrett2021integrated}. The robot must first search for a sequence of single joint changes $\Pi = [\theta_1, \ldots, \theta_T]$ using the scene level articulation model $M$, resulting in a geometrically feasible sequence of object motions that accomplishes the scene goal configuration $\Theta_g$. For each interaction, the robot must then infer a collision-free trajectory $\tau_i$ in the state space $\Theta \times S_R$ that accomplishes the subgoal $\theta_i$, including navigating to the object and manipulating it. If no such motion can be found (\eg articulating one object blocks the robot's path to a second), the planner should return to the high level and seek a new object-level plan. This process should be repeated until the robot finds a task plan $\Pi$ and corresponding collision-free motion plans $\{\tau_i\}_{i=1}^{T}$ that achieve the scene goal configuration $\Theta_g$.

\begin{figure*}[t]
    \centering
    \includegraphics[width=1.0\textwidth]{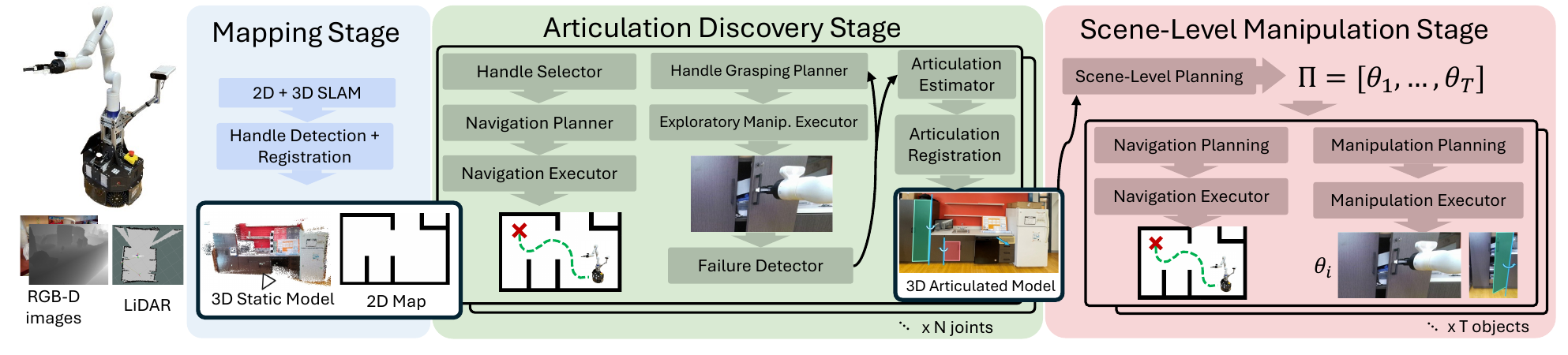}
    \caption{
        \textbf{\ourtool{} System Overview.} 
        Our approach involves three stages: a mapping stage (left), where the robot conducts a 3D scan and detects handles; an articulation discovery stage (middle), where the robot navigates, interacts, collects observations, and estimates the scene-level articulation model; and a scene-level manipulation stage (right), where the robot plans tasks using the scene-level articulation model and executes long-horizon actions through the articulation planner.
    }
    \vspace{-2mm}
    \label{fig: framework}
\end{figure*}

\section{Related Work}
In this section, we delve into previous research efforts related to the subproblems associated with scene-level articulated object manipulation.




\noindent \textbf{Articulation Model Estimation.}
Articulation model estimation is the problem of inferring the parts of articulated objects and the kinematic constraints between these parts. Objects are typically represented as a \textit{kinematic graphs} wherein nodes $V$ correspond to object parts and edges $E$ represent motion constraints between the parts.
Previous approaches utilized probabilistic methods~\cite{dearden2005learning, sturm2008adaptive, sturm2008unsupervised, sturm2009learning, sturm2011probabilistic} to infer these constraints through part tracking. However, their practical applicability in natural environments is limited due to the requirement of markers or handcrafted features. Leveraging large-scale 3D datasets~\cite{wang2019shape2motion, mao2022multiscan, gupta2023predicting}, recent methods have employed learning-based approaches for articulation estimation from raw sensory input including images~\cite{zeng2021visual, abbatematteo2019learning, gupta2024opening}, videos~\cite{qian2022understanding}, and point clouds~\cite{li2020category, wang2019shape2motion, weng2021captra, noguchi2021neural, hameed2022learning}. These models exhibit limited generalization to arbitrary unseen objects as they heavily rely on category-level priors for articulation parameter estimation. Recent studies have proposed category-independent methods~\cite{jain2020screwnet, jiang2022opd, sun2023opdmulti, heppert2022category, jiang2022ditto} to alleviate dependence on object categories. These methods typically require observing part motion in order to reliably infer the articulation models, necessitating human demonstrations or exploration strategies not addressed in these works. 

\noindent \textbf{Articulated Object Manipulation.}
Manipulating articulated objects requires coordinating the movements of a robot to ensure that the action sequence obeys the (unknown a priori) kinematic constraints of the objects. 
Previous studies suggest explicitly estimating articulation parameters through exploratory actions~\cite{katz2008manipulating, barragan2014interactive, wang2022adaafford, nie2023structure, wu2024learning, curtis2023task, martin2022coupled, gadre2021act}. The estimated model then facilitates manipulation, either through scripted actions~\cite{jiang2022ditto, curtis2023task}, planning~\cite{katz2014interactive, barragan2014interactive}, or learned policies~\cite{kumar2019estimating, lv2022sagci}. However, these studies are limited to controlled laboratory settings, primarily focusing on modeling a single object in a tabletop environment. 

More recent research investigates inferring affordances and applying actions based on initial observations~\cite{mo2021where2act, xu2022universal, wu2021vat, gupta2023predicting, xiong2024adaptive} or per-timestep~\cite{eisner2022flowbot3d, zhang2023flowbot++}.  However, a single initial observation remains inherently ambiguous in realistic scenarios as it does not uniquely determine the underlying kinematic constraints. These methods fail to extract object geometry or kinematics and therefore lack the capability for long-horizon planning.

\noindent \textbf{Integrated Systems for Scene-level Manipulation.}
Integrated systems for scene-level articulated object manipulation should possess capabilities for navigation, physical exploration, articulation model estimation, and task and motion planning. Earlier studies~\cite{meeussen2010autonomous, chitta2010planning} have primarily developed systems for door opening but fail to consider other articulation types. Task and motion planning approaches~\cite{stilman2007task, garrett2021integrated} typically assume accurate kinematic and geometric models and study only the planning aspect. 
Recent research~\cite{burget2013whole, arduengo2021robust, mittal2022articulated, gupta2024opening, xiong2024adaptive} introduces frameworks for interacting with unknown doors and drawers in indoor spaces. However, this research primarily focuses on single-object scenarios and, as a result, falls short when extending to a scene-level context for task-level planning. The most closely related work~\cite{hsu2023ditto} aims to build an articulation model of an indoor environment through a robot's self-directed exploration but is mainly focused on building digital twins without real-world downstream task evaluation. Instead, in this work, we develop a complete system for autonomously constructing and planning with a scene-level articulated model.






\section{\ourtool{}}

Our system to build and use scene-level articulated models operates in three stages (Fig.~\ref{fig: framework}):
in the initial \textit{mapping stage}, the robot builds a base model with 2D and 3D maps of the scene. The 3D map is enriched with a set of detected handles that guide the physical exploration in the next stage.
In the second \textit{articulation discovery stage}, the robot aims at finding and characterizing all joints in the scene. To accurately perceive that, the robot must generate motion actuating each degree of freedom~\cite{martin2014online,katz2013interactive,hsu2023ditto}. One by one, the robot moves to each detected handle and attempts to explore it reactively. With the observations from the interaction, it infers the possible articulation that is then registered into the scene-level 3D map for planning.
In the final \textit{scene-level manipulation  stage}, the robot utilizes the reconstructed scene-level articulation model to plan and execute more efficiently a change in the environment that requires actuating multiple DoF in the right sequence.
In the following, we provide details of each stage.


%

\begin{figure}
\removelatexerror
\renewcommand\footnoterule{}
\begin{algorithm}[H]
\caption{Articulation Discovery Stage}
\label{alg: exploration}

\SetNoFillComment
\SetKwInOut{Input}{Input}
\SetKwInOut{Output}{Output}
\SetKwFunction{GlobalNavTo}{GlobalNavTo}
\SetKwFunction{LocalNavTo}{LocalNavTo}
\SetKwFunction{PlanBase}{PlanBase}
\SetKwFunction{GetCurrentObs}{GetCurrentObs}
\SetKwFunction{DetectHandle}{DetectHandle}
\SetKwFunction{GetComplianceAction}{GetComplianceAction}
\SetKwFunction{ApplyAction}{ApplyAction}
\SetKwFunction{ArmRetract}{ArmRetract}
\SetKwFunction{DetectFailure}{DetectFailure}
\SetKwFunction{PredArticulation}{PredArticulation}
\SetKwFunction{Register}{Register}
\SetKwFunction{Reposition}{Reposition}

\Input{Handle location set $H$ and 3D static map $B$}
\Output{Scene-level Articulation Model $M$}
\tcc{initialize set of object models}
$A = \emptyset$\;

\ForEach{$h \in H$}{ 
    $b \leftarrow$ \PlanBase{$h$}\;
    \GlobalNavTo{$b$}\;
        $o \leftarrow$ \GetCurrentObs{}\;
        $p \leftarrow$ \DetectHandle{o}\;
        $b_{pre} \leftarrow$ \PlanBase{p}\;

    \Repeat{$f' = \emptyset$}{
        \LocalNavTo{$b_{pre}$}\;
        
        \tcc{get pre-action observation}
        $o_{pre} \leftarrow$ \GetCurrentObs{}\;

        \tcc{exploratory interaction}
        \For{$i\leftarrow 1$ \KwTo $n$}{
            $o' \leftarrow$ \GetCurrentObs{}\;
            $f' \leftarrow$ \DetectFailure{$o_{pre}$, $o'$}\;
            \If{$f' \neq \emptyset$}
                {$b_{pre} \leftarrow$ \Reposition{$o_{pre}$, $o'$} \;
                break}
            $a' \leftarrow$ \GetComplianceAction{o'}\;
            \ApplyAction{a'}\;
        }
                
    }
    $b_{post} \leftarrow$ \PlanBase{p}\;
    \LocalNavTo{$b_{post}$}\;
    \tcc{get post-action observation}
    $o_{post} \leftarrow$ \GetCurrentObs{}\;
    \tcc{predict articulation model}        
    $\alpha \leftarrow$ \PredArticulation{$o_{pre}$, $o_{post}$}\;
    $A \leftarrow A \cup \alpha$\;
}

$M \leftarrow$ \Register{A, B}\;
\end{algorithm}
\vspace{-8mm}
\end{figure}





\subsection{Mapping Stage}
In the mapping stage, our focus is on building the static model and identifying potential interaction regions (Figure~\ref{fig: framework}, left). Specifically, we employ SLAM to build the static base map $B$, and a handle detector to identify potential interaction hotspots for articulation exploration. 

\noindent \textbf{2D+3D SLAM.}
To build the static map, we manually navigated the robot around the scene and created a 3D scan with DROID-SLAM~\cite{teed2021droid}. We use the Nautilus~\cite{nautilus} tool based on the scene scan for 2D navigation mapping.

\noindent \textbf{Handle Detection and Registration.}
We are dealing with articulated objects designed for human use; therefore, we assume they incorporate handles that facilitate human-hand interaction. 
Consequently, we simplify the problem of finding interaction hotspots by focusing on handle detection. We use a YOLO detector~\cite{redmon2016you} trained on the DoorDetect dataset~\cite{arduengo2021robust} to detect handles. Given an image, the detector outputs the bounding box of the detected handle. We then use non-maximum suppression to remove duplicate handle detections and register the remaining to the global map.


\subsection{Articulation Discovery Stage}
In the articulation discovery stage (Figure~\ref{fig: framework}, middle), our focus is on physically interacting with the objects, creating informative observations, and inferring the scene-level articulation model. 

Interacting with the objects in this phase poses challenges due to the lack of prior understanding about the kinematic constraints. 
The robot must determine a base position and arm trajectory that accomplishes each manipulation without self-collision or joint limit violations. 
To overcome these challenges, we propose a progressive exploration strategy and failure recovery mechanism shown in Algorithm~\ref{alg: exploration}.

\noindent \textbf{Handle Selector.}
The robot first selects a candidate handle $h\in H$ to explore from the set of detected handle locations. 

\noindent \textbf{Navigation Planner and Executor.}
Upon identifying the candidate of interest on the map, the robot can navigate directly to the object (\verb|PlanBase, GlobalNavTo|). We use ENML~\cite{biswas2016episodic} for planning and localization during the navigation. 

Global localization is prohibitively inaccurate for object manipulation. To precisely position the base for manipulation, we use the MOSSE object tracker~\cite{bolme2010visual} to track the relative pose in the end-effector frame between the detected handle and the robot from the video stream. We implement a simple PID controller for the base movement to ensure the robot achieves the desired position (\verb|LocalNavTo|).

\noindent \textbf{Handle Grasping Planner.}
Based on the robot's egocentric observation, we use the handle detector to retrieve the bounding box of the handle. The center point of the bounding box is selected as the interaction hotspot. This location is then converted into the end effector frame by leveraging depth sensor information and the robot executes a grasp. 

\begin{figure}[t]
    \centering
    \includegraphics[width=\linewidth]{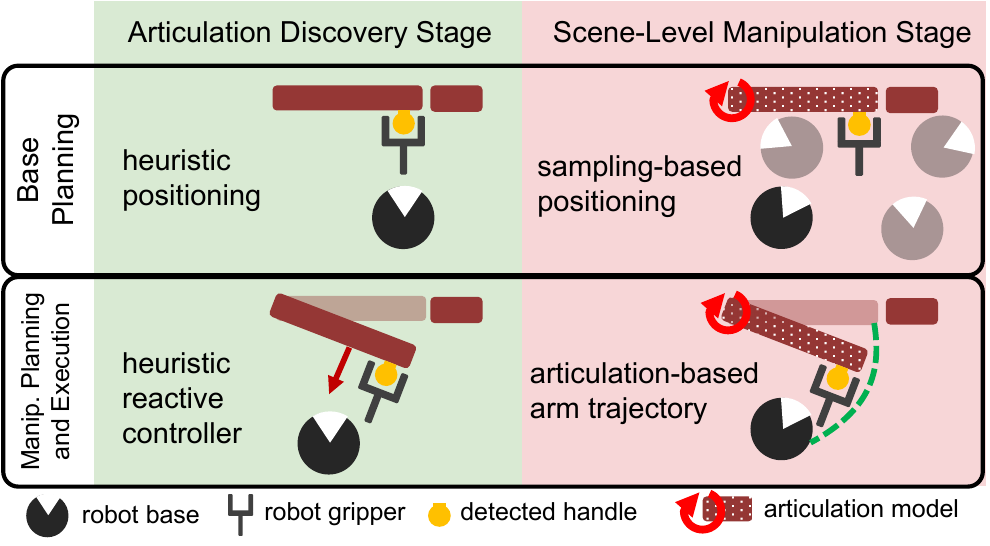}
    \vspace{-5mm}
    \caption{
        \textbf{Interaction through autonomous exploration vs. articulation planning.} In the Articulation Discovery stage (green), the agent does not have information about the articulation and performs heuristic-based base positioning (in front of the interaction point) and manipulation (admittance controller with motion normal to the plane). In the Scene-Level Manipulation Stage (light red), the robot plans an efficient base positioning and an arm trajectory based on the estimated articulation model obtained during exploration. Thanks to the built model, \ourtool{} achieves higher manipulation success and interacts faster than the baselines (Sec.~\ref{ss:soam}).
    }
    \vspace{-2mm}
    \label{fig: exploration-deployment}
\end{figure}

\noindent \textbf{Exploratory Manipulation Executor.}
To ensure that the applied force on the object complies with kinematic constraints, we estimate the surface normals of the point cloud observation from the wrist camera at each time step and adjust the end-effector action to pull perpendicular to the normal (\verb|GetComplianceAction|, Figure~\ref{fig: exploration-deployment}, left). These surface normal estimates predict how the object part would move under articulated motion without relying on an articulation model. This action is executed through an admittance controller to prevent damage to hardware or furniture (\verb|ApplyAction|).

\noindent \textbf{Failure Detector.}
Without knowledge of the articulation model, the robot's exploration attempts often fail due to sub-optimal base positions, self-collisions and joint limits. 
Interaction failures may also arise from inaccurate or absent depth information or the gripper slipping away during the interaction. To identify and address these errors, we incorporate a straightforward error detection mechanism, denoted \verb|DetectFailure| in Algorithm~\ref{alg: exploration}. By computing the distance between the observed point cloud before and after the interaction, we can determine whether the mobile part has been displaced by setting a certain threshold.

Upon detecting a failure, the robot repositions itself and makes another attempt at manipulating the object (\verb|Reposition|). One insight is that even if the manipulation attempt fails, it can provide crucial information about the joint type. For joint classification, we utilize plane detection on the wrist camera that can track the pose changes of the object's mobile part. Depending on the presence of rotation, we classify the joint as either revolute or prismatic. If rotation is detected, whether clockwise or counterclockwise, we further distinguish the rotation direction as left or right revolute. While the joint type cannot replace the need for a complete articulation model, this information can be leveraged to adjust the base for collecting better motion observations. For revolute joints, the robot moves to the same side as the rotation to preserve more space for subsequent manipulation. In the case of prismatic joints, the robot moves away from the object to allocate more space for the translation movements. Considering the workspace of the arm, the new base position is set within a fixed distance,~\ie, 30 cm, away from the interaction hotspot. It's important to note that these strategies do not account for object size and position, resulting in sub-optimal manipulation results. Nonetheless, they play a crucial role in the articulation model-building process by providing motion signals beyond static observations.

\noindent \textbf{Articulation Estimator.}
During exploration, the robot collects egocentric observations of the object before and after the interactions. By analyzing these observations, \ourtool{} can accurately predict the kinematics of the object by using a neural network to infer the articulation model, building upon the prior work Ditto in the House (DiTH)~\cite{hsu2023ditto}.

Visual occlusion poses a significant challenge when estimating object geometry and kinematics, especially when the camera is often obstructed by the robot arm or object parts during the manipulation of articulated objects. 
To address this challenge, we collect observations only before and after interactions. In practice, the robot will move a small fixed distance away from the object to capture observation $o_{post}$ (Algorithm~\ref{alg: exploration} lines 24-27). Empirically, we have observed that this improves the observation coverage and results in more accurate articulation estimation.

For each interaction, the robot acquires observation point clouds $o_{pre}$ and $o_{post}$ and interaction hotspots $c_{pre}$ and $c_{post}$. To integrate knowledge from the interaction, we incorporate the contact regions into our network input by creating Gaussian heatmaps centered around $c_{pre}$ and $c_{post}$ over the point clouds $o_{pre}$ and $o_{post}$, respectively. These heatmaps represent the contact regions of interaction and offer insights into the mobile part region and underlying kinematic constraints. Both the point cloud observations and the heatmaps of contact regions are fed into the network for articulation inference (\verb|PredArticulation|). The model then outputs the articulation joint parameters and part segmentation of each articulated object. 

\noindent \textbf{Articulation Registration.}
Finally, to construct the scene-level articulation model $M$, we register each estimated object model to the static base map $B$ using Colored ICP~\cite{park2017colored}, discarding outlier points (\verb|Register|). 

\subsection{Scene-Level Manipulation Stage}
In the scene-level manipulation stage, the robot utilizes the reconstructed scene-level articulation model to plan and execute sequential manipulation tasks. The key components for this stage are presented in the rightmost section of Figure~\ref{fig: framework}.

\noindent \textbf{Scene-Level Planning.} 
To successfully achieve the goal scene configuration $\Theta_g$, the robot must first determine a feasible sequence of object interactions, avoiding collisions between object parts or blocking its path to other objects. 
\ourtool{} first searches for a sequence of single joint changes $\Pi = [\theta_1, ..., \theta_T]$ such that each $\theta_i$ is feasible and the desired scene configuration $\Theta_g$ is achieved using a simple task and motion planning implementation~\cite{garrett2021integrated}. 

\ourtool{} begins by sampling object part trajectories for each object that must be manipulated to achieve $\Theta_g$. Bounding boxes for each part are extracted using the segmentation of the predicted articulation model, and the pose of each mobile part along the articulation trajectory is computed based on articulation parameters. In practice, we sample 6 configurations for each object by interpolating from zero to the maximum state (90 degrees for revolute joints and 15cm for prismatic). The order of interactions is then determined by sampling candidate plans and checking for feasibility. If the bounding boxes of the manipulated parts overlap or the path of the robot is blocked, the plan is ruled infeasible. \ourtool{} continues sampling plans until we find a feasible sequence of interactions. The robot then executes each interaction using the Manipulation Planner described below to achieve the scene goal configuration $\Theta_g$. 



\noindent \textbf{Manipulation Planning and Execution.}
At each interaction, the robot uses the inferred articulation models to manipulate the objects. The robot grasps the interaction hotspot $p \in \mathbb{R}^{3}$ and applies actions that comply with the kinematic constraints until reaching the goal angle $g_r$ for revolute joint, or goal translation $g_p$ for prismatic joint. These sequential actions can be represented as trajectories following prior work~\cite{eisner2022flowbot3d, zhang2023flowbot++}. To determine a base position that enables manipulation (preventing self-collision or reaching joint limits), we use a random sampling approach. We sample a set of base positions within a specified range around the object. Among these base positions, we choose the one that can reach the most poses on the trajectory as our final base position during execution (Figure~\ref{fig: exploration-deployment}, right).

For a revolute joint, the action trajectory lies within the circle within a plane perpendicular to the revolute axis $u^r$ with origin $q$. Given a rotation angle $\phi$, the rotation matrix can be defined as follows:
\begin{equation}
    \mathbf{R}(\phi) = \mathbf{I} + \sin \phi [u^r]_\times + (1-\cos\phi) [u^r]_\times^2
    \label{eq:get-rot}
\end{equation}
where $\mathbf{I}$ is identity matrix and $[u^r]_\times$ is the skew-symmetric matrix of $u^r$.
By evenly sampling $K$ steps between $0$ and the goal angle $g_r$, the action trajectory can be defined as follows:
\begin{equation}
    \tau_\text{revolute} =\left\{\mathbf{R}\left(\frac{i}{K}{g_r}\right)(p-q)+q\right\}_{\forall i \in[0, K]}
    \label{revolute-traj}
\end{equation}

For prismatic joints, the $K$-step trajectory is defined as the translation of the grasp point $p$ to the goal translation $g_p$ along the translation axis $u^p$.
\begin{equation}
    \tau_\text{prismatic} = \left\{p + \frac{i}{K}g_p {u^p} \right\}_{ i \in[0, K]}
    \label{prismatic-traj}
\end{equation}
\noindent
The action trajectory is then executed with a position controller.

In contrast to the articulation discovery stage, the complete trajectory is now available, eliminating the need to recalculate each step during the actuation of the articulated part.
The end-effector trajectory is executed via a position controller to achieve the desired manipulation, and the robot proceeds to the subsequent interactions in the scene-level plan.

\section{Experiments}

We evaluate 
1) \ourtool{}'s effectiveness at performing  long-horizon tasks that involve scene-level articulated manipulation; 
2) the effectiveness of different controllers for single-object articulated manipulation compared to \ourtool{};
3) the accuracy of articulation estimation using different algorithms within \ourtool{}; and
4) the sources of error in the autonomous discovery of articulated objects using \ourtool{}.

\begin{figure}[t]
    \centering
    \includegraphics[width=\linewidth]{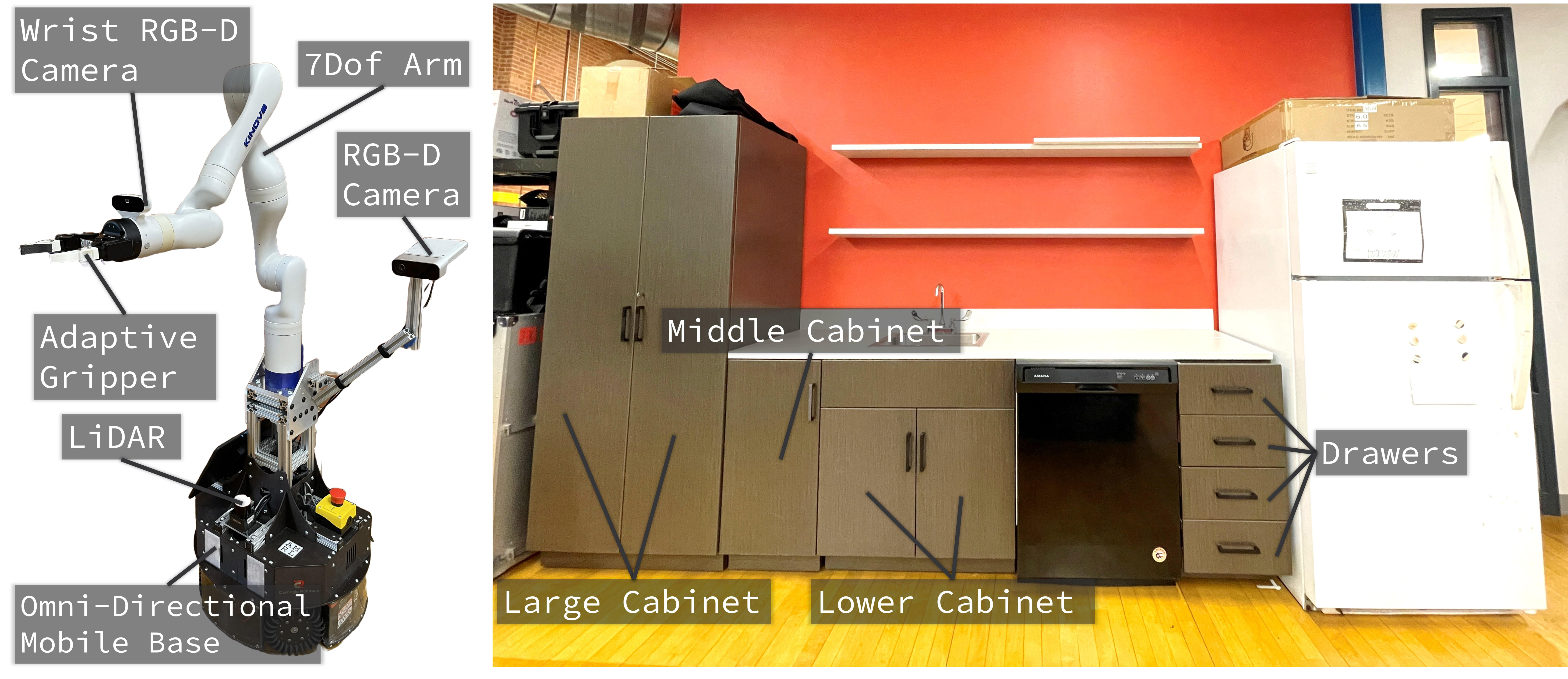}
    \vspace{-5mm}
    \caption{
        \textbf{Our mobile manipulator and the indoor kitchen scene.} (Left) We integrate a custom omnidirectional base with a 7 DoF torque-controlled arm and a 2-fingered hand, combined with two RGB-D and one LiDAR sensors. (Right) Our kitchen environment contains 7 degrees of freedom including revolute and prismatic in objects of different shapes, weights and heights.
    }
    \vspace{-2mm}
    \label{fig: hardware}
\end{figure}

We evaluate \ourtool{} in a real-world kitchen scene comprising 4 drawers and 5 cabinet doors of different sizes and shapes, shown in Figure~\ref{fig: hardware}. Our mobile manipulator robot is equipped with a four-wheeled custom omnidirectional base and a Kinova Gen3 7 DoF arm. 
The robot utilizes a forward-facing Kinect RGB-D camera, a Hokuyo UST-10LX 2D LIDAR for scene mapping and localization, and an Intel Realsense RGB-D camera on the arm wrist to guide manipulation. 
A Robotiq F-140 end-effector with 3D-printed soft robotic fingers offer flexibility and compliance during manipulation. 

\subsection{Scene-level Articulated Manipulation}
We generate five sequences wherein the robot must interact with three articulated objects of varying sizes, positions, and articulation properties within the scenes. Success in each task requires the robot to consider how the outcome of its actions would impact its future plans, similar to the scenario depicted in Fig.~\ref{fig: teaser}. 
Provided with the initial position and the objects requiring articulation over a minimum threshold, the robot autonomously navigates and infers an action sequence to actuate the objects. The entire process can be witnessed in the accompanying video. We compare our method with a \textit{Random} baseline that samples randomly a sequence of object manipulations and executes it. To ensure a fair comparison, we conduct five runs for each attempt for both methods.

\begin{figure}[t]
    \centering
    \includegraphics[width=\linewidth]{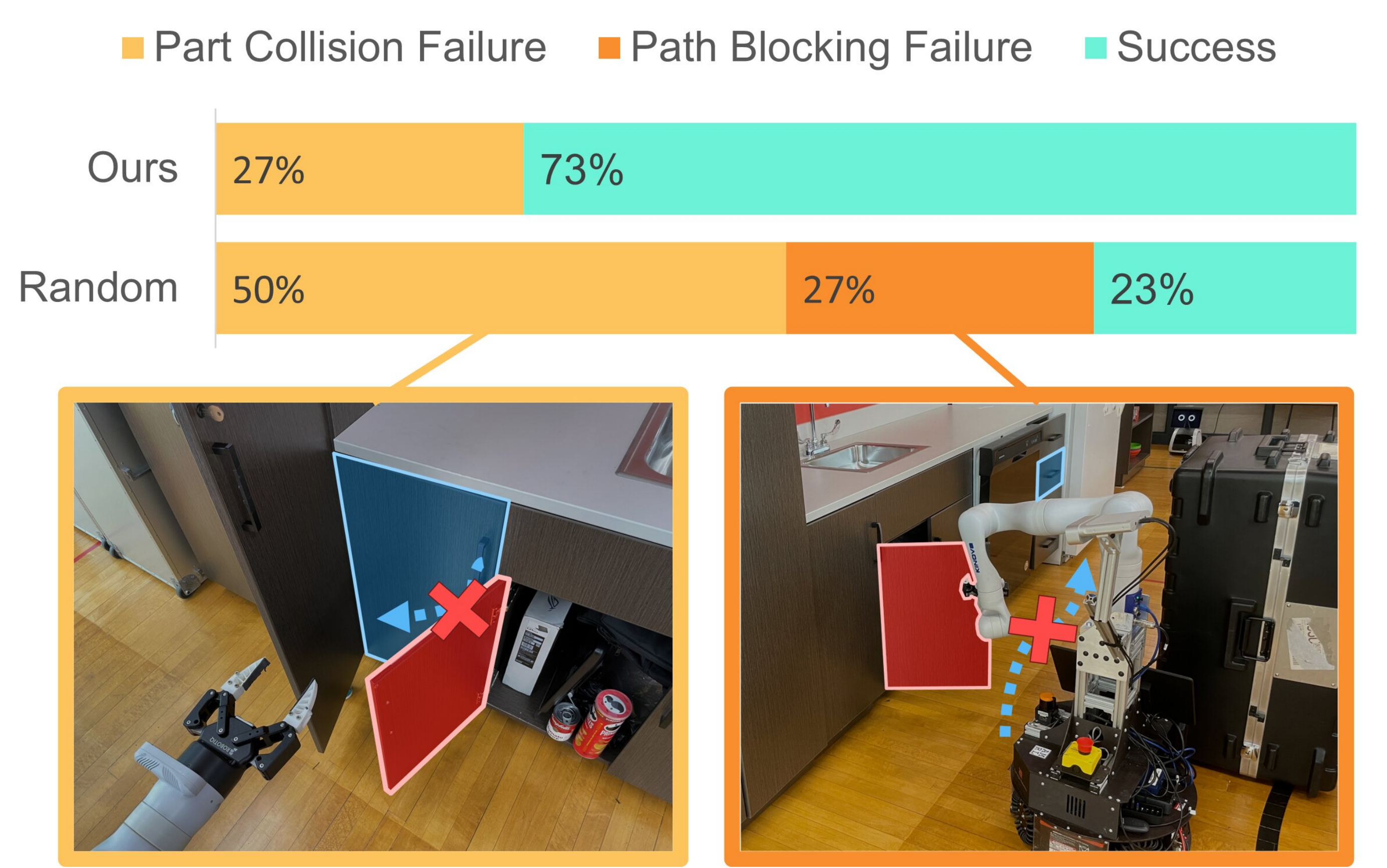}
    \vspace{-5mm}
    \caption{
        \textbf{Results of Scene-level Manipulation.} \ourtool{} enables planning sequences of articulated object interactions. The lower left figure illustrates an example of part collision where the rightmost cabinet blocks actuation of the middle cabinet. The lower right figure illustrates the path blocking failure, where actuating the cabinet first blocks the robot's path to other objects in the plan.
    }
    \vspace{-2mm}
    \label{fig: planning}
\end{figure}

Figure~\ref{fig: planning} tabulates the results --- we identify two failure modes, namely part collision failures (an object part cannot be actuated because it collides with another), and path blocking failures (the path of the robot is blocked by an object's part). The Random baseline exhibits a mere 23\% success rate and suffers significantly from both object part collisions and mobility blocking failures. 
In contrast, our method achieves a 73\% success rate with 23\% fewer object part collision failures and no path blocking failures. The occasional failure of our method is due to erroneous articulation estimation that results in a failure to detect a part collision. These results empirically demonstrate the effectiveness of \ourtool{}'s approach in unfamiliar scenes with exploration and modeling followed by scene-level model-based planning and control. 

\subsection{Single Object Articulated Manipulation}
\label{ss:soam}
We compare \ourtool{}'s model-based approach to state-of-the-art zero-shot articulated manipulation controllers to evaluate the importance of model-based articulation estimation on robust manipulation.
We report the execution time, opening degree, and execution success rate in single object manipulation. The execution time is determined by the duration between the initiation of the action and the point at which the object has been articulated over a minimum threshold of 30$^\circ$ for revolute joints and \SI{5}{\centi\meter} for prismatic joints. We define the \textit{opening degree} as the ratio of the actuation degree to the maximum degree. 

We compare \ourtool{} model-based manipulation strategy with three baselines: 1) \textbf{FlowBot3D~\cite{eisner2022flowbot3d}}, a zero-shot approach that generalizes the manipulation of articulated objects from a large amount of training data.
2) \textbf{Exploration-Only},  where the interaction uses only the heuristic strategy employed in our articulation discovery stage, without building articulation models.
3) \textbf{Model-Based Oracle},  where the interaction is conducted using \ourtool{}'s model-based approach but assuming ground-truth knowledge of the articulation model. It represents an upper-bound scenario achievable for our model-based approach.

Table~\ref{tab: manipulation} tabulates the execution results. For both revolute and prismatic joints, \ssfont{FlowBot3D} is unable to articulate any object to the required threshold. Two factors contribute to this failure: first, \ssfont{FlowBot3D} is trained and tested in a controlled lab environment where the optimal viewing angle is manually configured and ensured. And second, the presence of a cluttered environment, and variations in object shape, and size introduce a considerable domain gap, leading to the observed shortcomings. On the other hand, \ssfont{Exploration-Only} effectively articulates all objects to the threshold, gathering observations used to construct the articulation model, but it requires iterative updates to its base position due to self-collision and reaching joint limitations before succeeding, adding execution time. Moreover, it only infers per-step actions throughout the interaction, also contributing to a slower execution speed.

In contrast, \ourtool{} (\ssfont{Ours}) strategically plans the base position using pre-computed action trajectories derived from the acquired articulation model. As a result, \ssfont{Ours} achieves a base reposition speed that is 2.88 times faster and executes arm actions 3.04 times faster for revolute joints and 2.17 times faster for prismatic joints compared to \ssfont{Exploration-Only}. It successfully reaches 78\% of the maximum opening degree compared to \ssfont{Model-based Oracle}. This gap suggests that \ourtool{} can be further improved by more advanced articulation estimation methods in the future.


\begin{table}[h]
    \scriptsize
    \centering
    \vspace{6mm}
    \caption{Results of Single Object Manipulation.}
    \resizebox{\columnwidth}{!}{%
    \begin{tabular}{l|ccc|cc}
    \toprule
\raisebox{\dimexpr-1\normalbaselineskip-2\cmidrulewidth-2\aboverulesep}[0pt][0pt]{Method} & \multicolumn{3}{c|}{Execution time} & \multicolumn{2}{c}{Opening Degree} \\
      \cmidrule{2-6}
      & Base\,$\downarrow$ & Arm (Rev.)\,$\downarrow$ & Arm (Pris.)\,$\downarrow$ & Rev.\,$\uparrow$ & Pris.\,$\uparrow$ \\
     \midrule
     FlowBot3D~\cite{eisner2022flowbot3d} & - & - & - & 0.13 & 0.23 \\
     Exploration-Only & 9.85 & 10.37 & 9.57 & 0.43 & 0.92 \\
     Ours & 3.41 & 3.41 & 4.40 & 0.78 & 1.00 \\
     Oracle & 3.39 & 2.99 & 4.68 & 1.00 & 1.00 \\
     \bottomrule
    \end{tabular}
    }
    \label{tab: manipulation}
    \vspace{-5mm}
\end{table}

\subsection{Accuracy of Articulation Estimation}
We investigate several state-of-the-art methods for articulation estimation in \ourtool{}, and identify key factors affecting their performance. We use the same metrics as in prior work~\cite{hsu2023ditto}: axis orientation error (Angle Err.) for prismatic and revolute joints, and additionally axis position error (Trans Err.) for revolute joints. 

Table~\ref{tab: articulation} lists the articulation estimation errors in our real-world evaluation scene, over all articulations present in the scene. \ssfont{Art3D} relies on RGBD sequence input and thus performs poorly due to the occlusions and imperfect view angles that occur during object interaction. \ssfont{Proprioception} performs estimation based on the sequential end-effector pose collected during the physical interaction. This baseline outperforms the vision-based baselines but fails to recover object geometry or part segmentations necessary for long-horizon planning. 
\ssfont{OPDMulti} takes only the initial RGB observation of the object without observing part motion, and thus produces erroneous results. In contrast, DiTH~\cite{hsu2023ditto} (Ours) takes the observation before and after the interaction and accurately infers object kinematics, significantly outperforming all baselines.

\begin{table}[h]
    \scriptsize
    \centering
    \vspace{3mm}
    \caption{Quantitative results of articulation estimation.}
    \resizebox{\columnwidth}{!}{%
    \begin{tabular}{ll|c|cc}
    \toprule
\multicolumn{2}{c|}{Method} & \multicolumn{3}{c}{Joint} \\
      \cmidrule{1-5}
      \raisebox{\dimexpr-1\normalbaselineskip-1\cmidrulewidth-1\aboverulesep}[0pt][0pt]{\shortstack[c]{Name}} & \raisebox{\dimexpr-1\normalbaselineskip-1\cmidrulewidth-1\aboverulesep}[0pt][0pt]{\shortstack[c]{Input \\ Modalities}} & Prismatic & \multicolumn{2}{c}{Revolute} \\
      \cmidrule{3-5}
      & & Angle Err.\,$\downarrow$ & Angle Err.\,$\downarrow$ & Trans Err.\,$\downarrow$ \\
     \midrule
     Art3D\footnotemark{}~\cite{qian2022understanding} & RBG-D Seq. & 62.50 & 46.02 & 0.29 \\
     Proprioception & Pose Seq. & 2.77 & 5.43 & 0.19 \\
     OPDMulti~\cite{sun2023opdmulti} & RGB-D & 73.21 & 11.48 & 0.78 \\
     DiTH (Ours)~\cite{hsu2023ditto} & Two Point Clouds & 8.11 & 4.07 & 0.08 \\
     \bottomrule
     \multicolumn{5}{l}{\footnotesize $^{1}$ The method fails to detect articulation in 4 out of 8 sequences}\\
    \end{tabular}
    }
    \label{tab: articulation}
    \vspace{-5mm}
\end{table}

\subsection{Analysing Errors During Articulation Discovery}
\begin{figure}
 \centering
 \begin{subfigure}[b]{0.49\linewidth}
     \centering
     \includegraphics[width=\textwidth]{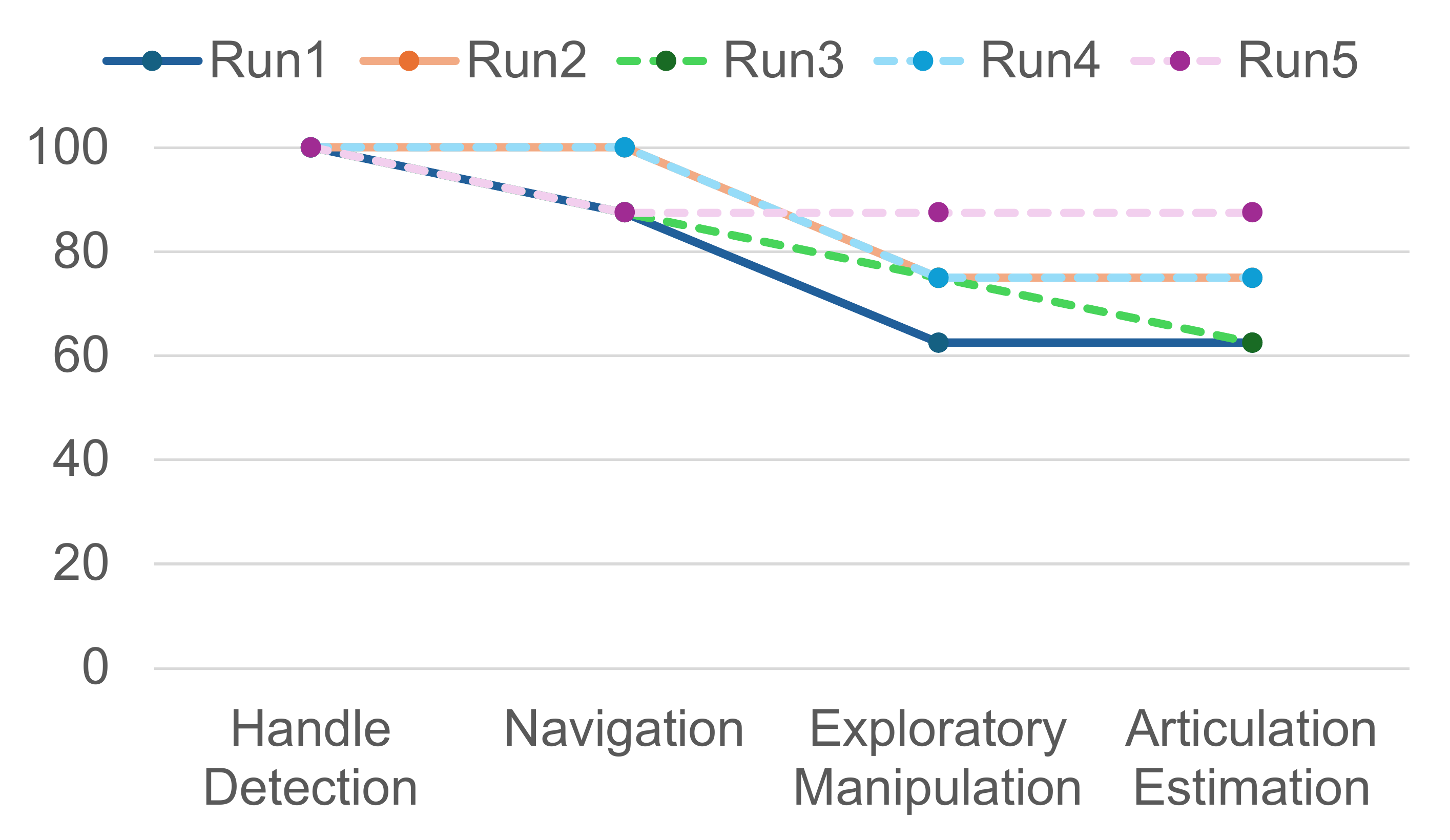}
     \caption{Success rate (in $\%$) for each trial after completion of each step}
     \label{fig:exploration-runs}
 \end{subfigure}
 \hfill
 \begin{subfigure}[b]{0.49\linewidth}
     \centering
     \includegraphics[width=\textwidth]{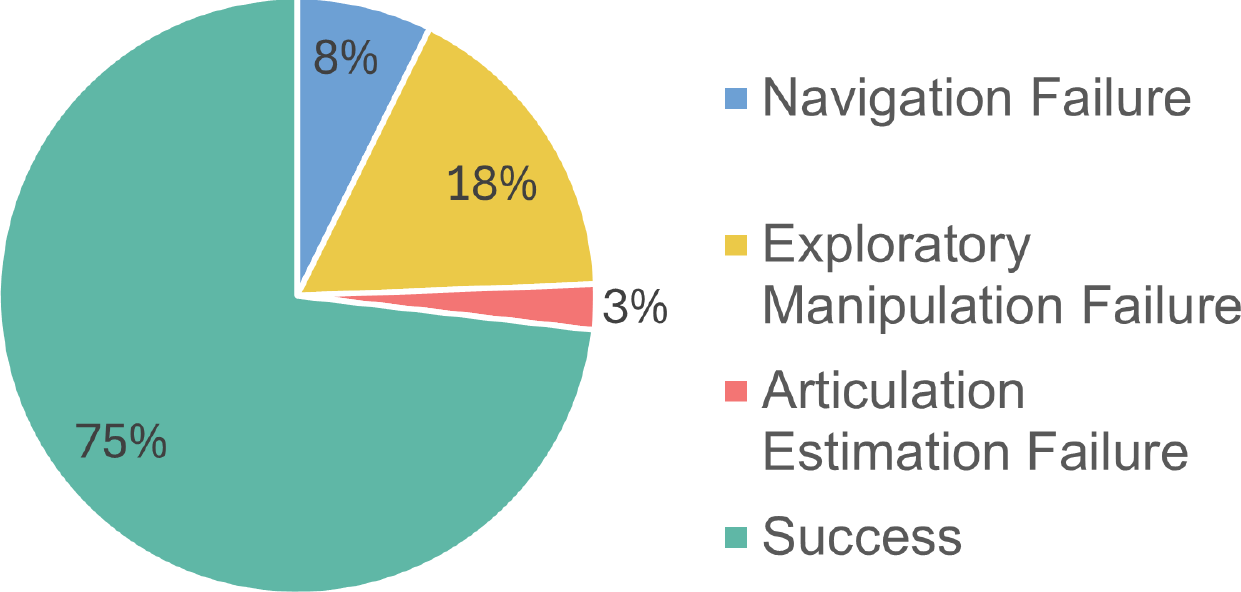}
     \caption{Cumulative success and sources of failures over all trials}
     \label{fig:exploration-pie}
 \end{subfigure}
\vspace{0.3em} 
\caption{\textbf{Analysing errors during articulation discovery}}
        \label{fig:exploration}
\vspace{-2mm}
\end{figure}

We investigate the sources of errors during articulation discovery since \ourtool{} relies on the resulting articulation models to successfully perform scene-level articulation tasks during deployment.
Figure~\ref{fig:exploration} shows the sources of errors over five repeated and independent runs of articulation discovery in the scene.
For each run, we logged (Figure~\ref{fig:exploration-runs}) the success rates over all articulated objects in terms of the percent of successful 1) detection of handles to articulated objects (Mapping), 2) navigation to the object based on heuristic positioning (Navigation), 3) zero-shot exploratory manipulation, and 4) articulation estimation. We also report the successes and failures over all runs and all objects (Figure~\ref{fig:exploration-pie}) in terms of the success at articulation discovery; and failures of each of the aforementioned four steps.
Typically, the main causes of failure stem from exploratory manipulation. These failures predominantly occur when our tracker loses track of the relative pose between the robot and the object during local navigation. Another factor contributing to failure is unreliable motion prediction, leading the arm to reach its limits and consequently fail to recover.


\section{Conclusion and Limitations} 
\label{sec:conclusion}

We introduced \ourtool{}, a method for model-based manipulation of articulated objects at the scene level, and showed experimentally that it can autonomously build an articulated 3D model of the scene, and use this model to plan and execute sequences of interactions.

Despite its success, \ourtool{} is not without limitations:  we assume that 1) the scene can be represented by a single-level kinematic tree; 2) the articulated objects have detectable and graspable handles, and 3) that the objects have plane normals perpendicular to the articulation motion. 
Despite these limitations, \ourtool{} is a promising initial solution to the problem of scene-level reasoning for articulation manipulation, which is necessary for robots to successfully perform long-horizon tasks in real human environments.


\renewcommand*{\bibfont}{\footnotesize}
\printbibliography 


\end{document}